\documentclass[letterpaper, 10 pt, conference]{ieeeconf}

\IEEEoverridecommandlockouts    % This command is only needed if 
\usepackage{times} % Postscript Times Font
\usepackage{graphicx} % Figure Support
\usepackage[ruled]{algorithm2e}  % Algorithm support
\usepackage[utf8]{inputenc} % UTF-8 Encoding support
\usepackage{setspace} % Change bibliography spacing
\usepackage[noadjust]{cite}

\usepackage{amsmath,amsfonts,amssymb,amsopn,bm}

\newcommand{\1}{\mathbf{1}}
\newcommand{\E}{\mathbb{E}}

\newcommand{\xxnote}[3]{}
\ifx\hidenotes\undefined
  \usepackage{color}
  \renewcommand{\xxnote}[3]{\color{#2}{#1: #3}}
\fi

\newcommand\figref{Figure~\ref}

\title{\LARGE \bf
Adaptive Robot-Assisted Feeding: An Online Learning \\
Framework for Acquiring Previously Unseen Food Items
}

\author{Ethan K. Gordon$^{1}$,
Xiang Meng$^{2}$,
Tapomayukh Bhattacharjee$^{1}$,
Matt Barnes$^{1}$ and
Siddhartha S. Srinivasa$^{1}$
\thanks{$^{1}$ Ethan K. Gordon, Tapomayukh Bhattacharjee, Matt Barnes and Siddhartha S. Srinivasa are with the Department of Computer Science and Engineering, University of Washington, Seattle, WA 98195
        {\tt\small \{ekgordon, tapo, mbarnes, siddh\}@cs.washington.edu}}%
\thanks{$^{2}$Xiang Meng is with the Department of Statistics, University of Washington, Seattle, WA 98195, 
        {\tt\small xmeng051@uw.edu}}%
}

\begin{document}

\maketitle
\thispagestyle{empty}
\pagestyle{empty}

%%%%%%%%%%%%%%%%%%%%%%%%%%%%%%%%%%%%%%%%%%%%%%%%%%%%%%%%%%%%%%%%%%%%%%%%%%%%%%%%
\begin{abstract}
A successful robot-assisted feeding system requires bite acquisition of a wide variety of food items. It must adapt to changing user food preferences under uncertain visual and physical environments. Different food items in different environmental conditions require different manipulation strategies for successful bite acquisition. Therefore, a key challenge is how to handle previously unseen food items with very different success rate distributions over strategy. Combining low-level controllers and planners into discrete action trajectories, we show that the problem can be represented using a linear contextual bandit setting. We construct a simulated environment using a doubly robust loss estimate from previously seen food items, which we use to tune the parameters of off-the-shelf contextual bandit algorithms. Finally, we demonstrate empirically on a robot-assisted feeding system that, even starting with a model trained on thousands of skewering attempts on dissimilar previously seen food items, $\epsilon$-greedy and LinUCB algorithms can quickly converge to the most successful manipulation strategy.

\end{abstract}

%%%%%%%%%%%%%%%%%%%%%%%%%%%%%%%%%%%%%%%%%%%%%%%%%%%%%%%%%%%%%%%%%%%%%%%%%%%%%%%%
\section{INTRODUCTION}
\label{sec:intro}
Eating is an activity of daily living that many of us take for granted. However, according to a U.S. study in 2010, approximately 1.0 million people need assistance to eat \cite{2012Brault}. The ability to self feed would not only save time for caregivers, but it would also increase a person's sense of self worth \cite{1990Prior, 1994Stanger}. Available commercial feeding systems \cite{myspoon, obi} have minimal autonomy and require preprogrammed movements, making it difficult for them to adapt to environmental changes. In general, a robust feeding system must be able to acquire a bite of food in an uncertain environment (``bite acquisition'') and transfer it safely to a potentially unpredictable user (``bite transfer''). Both are difficult and important problems, but this work focuses only on bite acquisition, and specifically the acquisition of food items that the robot may not have seen or manipulated before.

Different food items require different manipulation strategies for bite acquisition \cite{bhattacharjee2018food}. While recent work has achieved some successes in developing strategies that can acquire a variety of food items \cite{2019Feng, 2019Gallenberger}, it is unclear which strategy works best for previously unseen food. Even food items that look similar, such as ripe and un-ripe banana slices, can have very different consistencies, leading to different bite acquisition strategies. Our key insight is that we can leverage high-level successful bite acquisition strategies derived from human user studies \cite{bhattacharjee2018food} and an existing model batch-trained on a set of food items to suggest strategy success probabilities \cite{2019Feng} to perform online learning. 

We believe that exploring online learning bite acquisition can lead to manipulation strategies that better generalize to previously unseen food items. This is due to (a) the covariate shift from the training data set, (b) the diversity of food categories, and (c) the expensive process of collecting data on a physical robot. Factors that may contribute to covariate shift include changing lighting conditions, backgrounds, and not knowing the distribution of food items \emph{a priori}. An online learning scheme lets the system leverage data collected in real-world conditions and adapt to each user's specific palate.

\begin{figure}[t!]
    \centering
    \includegraphics[width=0.805\linewidth]{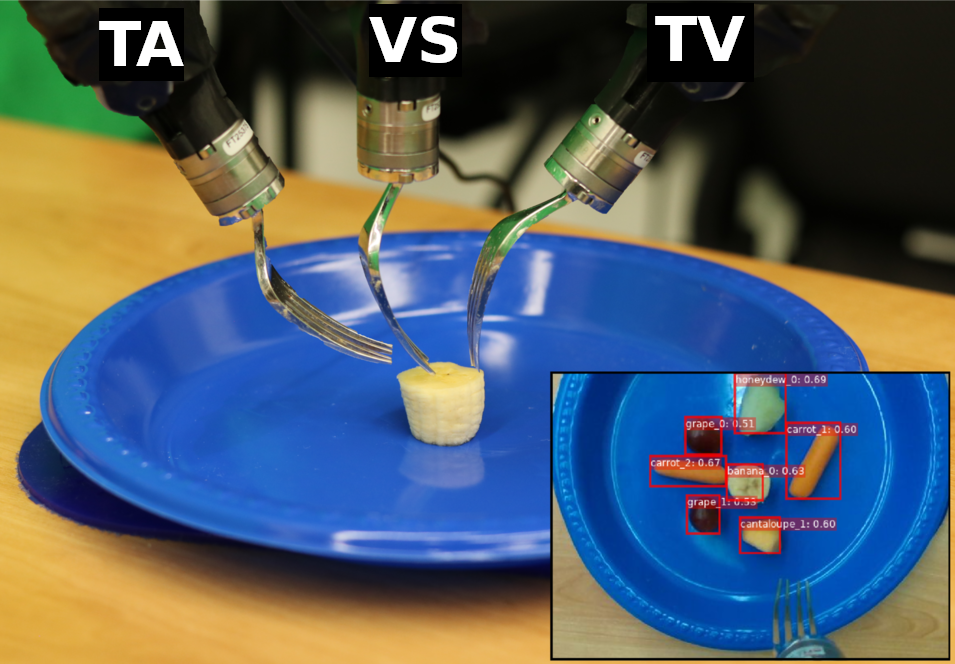}
    \includegraphics[width=0.175\linewidth]{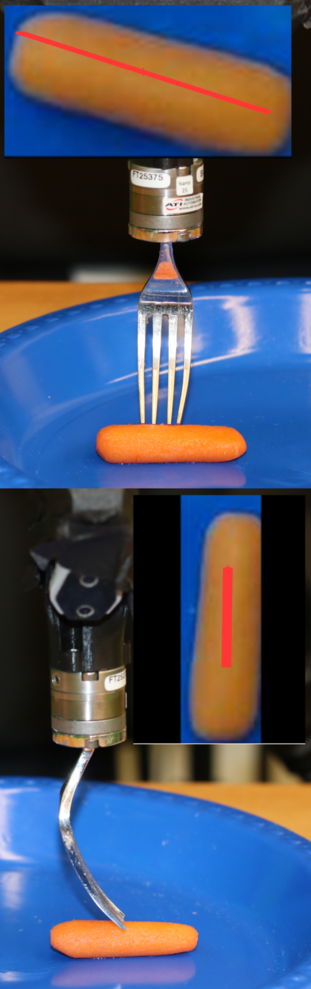}
    \caption{When faced with new foods, a robot-assisted feeding system must decide between a variety of bite acquisition strategies. In our system, each strategy is parameterized by a fork pitch (\emph{left}: tilted-angle ``TA", vertical ``VS", or tilted-vertical ``TV") and fork roll angle (\emph{right}: parallel or perpendicular). Our contextual bandit online learning framework learns from limited feedback (success or failure) after attempting each bite acquisition strategy, which itself depends on perception, planning, and low-level control.}
    \label{fig:overview}
    \vspace{-0.4cm}
\end{figure}

Importantly, each individual strategy returns only partial (or {\em bandit}) feedback. In other words, when our system takes an action to acquire a food item, it can see only whether it has failed or succeeded with that action. It is not privy to the counterfactual loss of other actions. Additionally, visual features provide context for each food item. Therefore, the problem naturally fits into the well-studied {\em contextual bandit} setting. 

In this work, we propose a contextual bandit framework for this problem setting. We present multiple algorithms based on the contextual bandit literature that could provide potential solutions. Our major contributions are (1) a framework, including a featurizer, simulated hyper-parameter tuner, and integrated off-the-shelf $\epsilon$-greedy \cite{Bietti2018} and LinUCB \cite{Chu2011} algorithms, and (2) empirical evidence of the framework's efficacy in real robot bite acquisition experiments. Our initial action space of 3 fork roll angles (tilted-angled (TA), vertical (VS), and tilted-vertical (TV), as shown in \figref{fig:overview}) $\times$ 2 fork pitch angles currently limit us to discrete, solid food items, but future work can examine a richer action space to tackle bite acquisition on even more varied food items and realistic plates.

%%%%%%%%%%%%%%%%%%%%%%%%%%%%%%%%%%%%%%%%%%%%%%%%%%%%%%%%%%%%%%%%%%%%%%%%%%%%%%%%

\begin{figure}[t!]
    \centering
    \includegraphics[width=\linewidth]{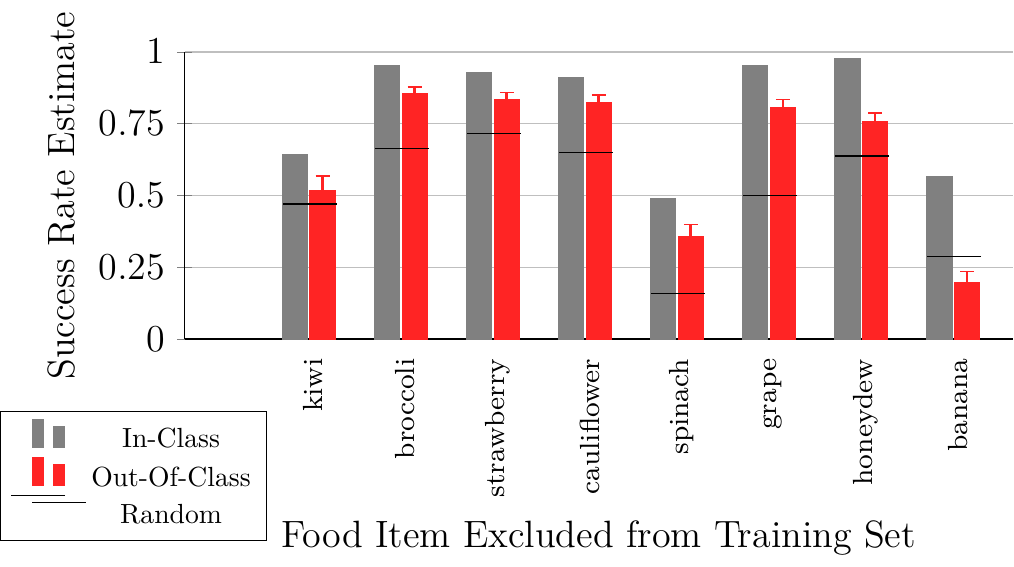}
    \caption{Generalization results for SPANet on select food items using data from \protect\cite{2019Feng} and the unbiased estimator described in Section \protect\ref{sec:dr}. When excluded from the training set, each item performs worse, with banana in particular performing significantly worse even than random ($p<0.05$).}
    \label{fig:spanet}
    \vspace{-0.5cm}
\end{figure}

\section{RELATED WORK}
\subsection{Robot-Assisted Feeding: Food Manipulation}
\label{sec:prev}
Food manipulation has been studied in various environments, such as the packaging industry~\cite{chua2003robotic, erzincanli1997meeting, morales2014soft, brett1991research, williams2001teaching, blanes2011technologies}, with focus on the design of application-specific grippers for robust sorting and pick-and-place, as well as showing the need for visual sensing for quality control~\cite{brosnan2002inspection, du2006learning, ding1994shape} and haptic sensing for grasping deformable food items without damaging them~\cite{chua2003robotic, erzincanli1997meeting, morales2014soft, brett1991research, williams2001teaching, blanes2011technologies}. Research labs have also explored meal preparation~\cite{ma2011chinese, sugiura2010cooking} as an exemplar multi-step manipulation problem, baking cookies~\cite{bollini2011bakebot}, making pancakes~\cite{beetz2011robotic}, separating Oreos~\cite{oreovideo}, and preparing meals~\cite{gemici2014learning} with robots. Most of these studies either interacted with a specific food item with a fixed manipulation strategy~\cite{bollini2011bakebot, beetz2011robotic} or used a set of food items for meal preparation that required a different set of manipulation strategies~\cite{gemici2014learning}.

Existing autonomous robot-assisted feeding systems~\cite{2019Feng, 2019Gallenberger, 2016Park, herlant_thesis} can acquire a fixed set of food items and feed people, but it is not clear whether these systems can adapt to very different food items that require completely different strategies. Feng {\em et al.} \cite{2019Feng} developed the {\em Skewering Position Action Network (SPANet)} and show generalization to previously unseen food items, but only for those with similar bite acquisition strategies. The universe of food items is massive; thus, it is almost impossible to train these systems on every kind of food items available. Even if we could, a static model is still vulnerable to the covariate shift (see Section \ref{sec:intro}). Our paper addresses this gap in the food manipulation literature by developing methods that can generalize to previously unseen food items with very different action distributions. We propose to use an online learning framework in a contextual bandit setting for food manipulation.

\begin{figure}[t!]
    \centering
    \includegraphics[width=\linewidth]{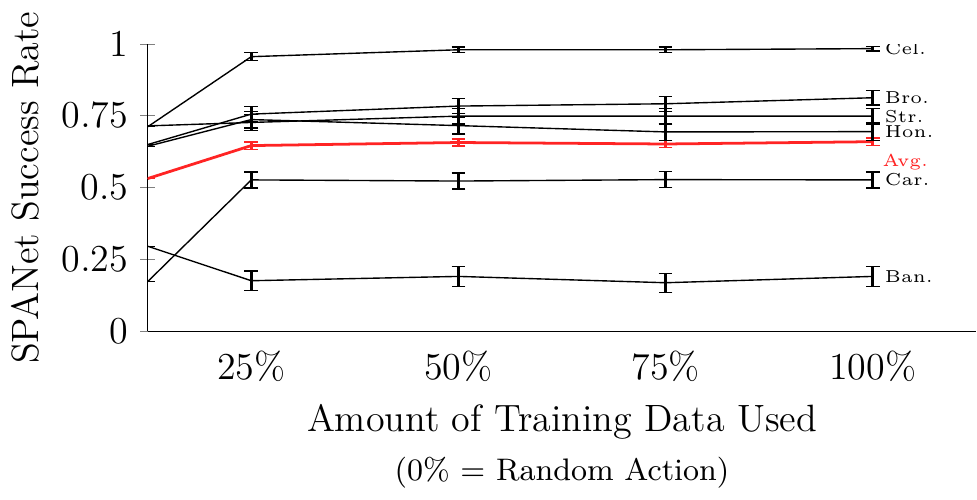}
    \includegraphics[width=\linewidth]{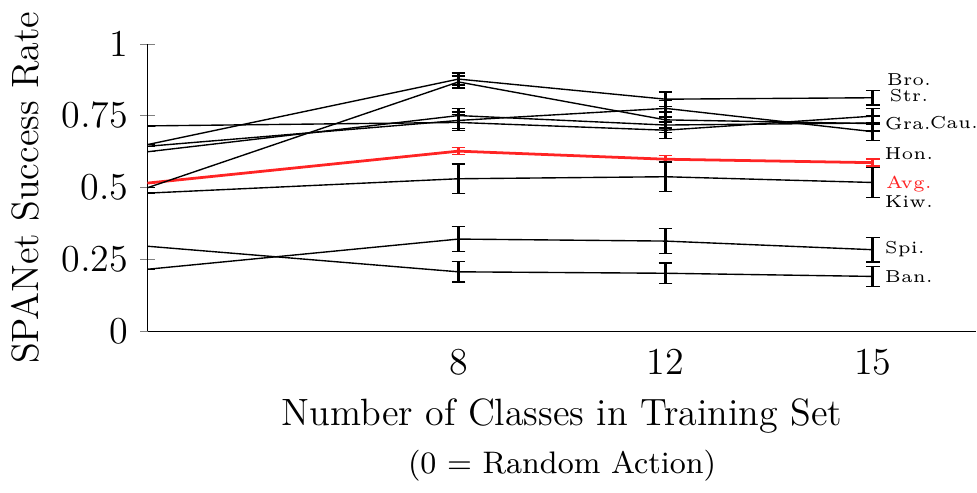}
    \caption{SPANet out-of-class success rate using data from \protect\cite{2019Feng}, given different amounts of data and food classes included in the training data set. Each black line represents a single food item excluded from the training set. The red line represents the performance averaged across all food items. The amount of out-of-class training data has already reached the point of diminishing returns at best. For very different food items (like banana), extra data actually reduces performance, likely due to over-fitting on the fixed set of food classes.}
    \label{fig:data_amount}
    \vspace{-0.3cm}
\end{figure}
\begin{figure*}[t!]
    \centering
    \includegraphics[width=0.95\linewidth]{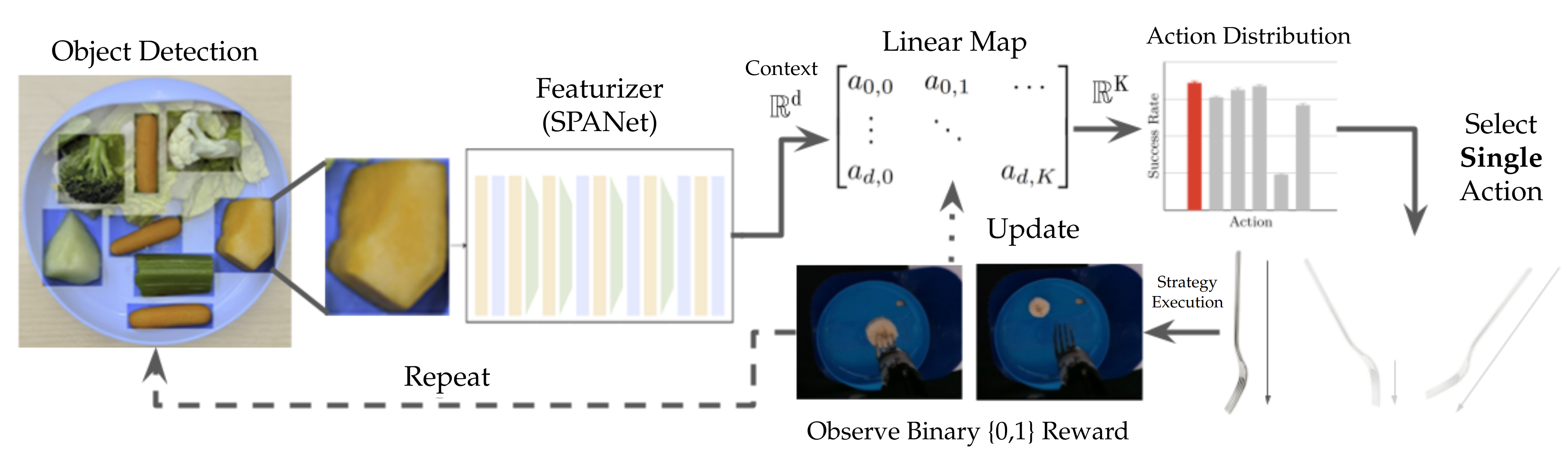}
    \caption{Online learning framework. SPANet is trained on previously seen food items, and then all but the last layer is frozen as a featurizer. The final linear layer becomes the ``linear map'' that we update after each subsequent attempt. The result is the estimated success rate of each action on the given food item, which we use to select a single action to attempt before updating the linear map.}
    \label{fig:system}
    \vspace{-0.3cm}
\end{figure*}
\subsection{Online Learning}
Bandit algorithms have seen widespread success in online advertising \cite{Tang2013,Bottou2013}, health interventions \cite{Klasnja2015,Hochberg2016}, clinical trials \cite{Shortreed2011}, adaptive routing \cite{Awerbuch2004}, education \cite{Mandel2014}, music recommendations \cite{Wang2014}, financial portfolio design \cite{Shen2015}, and any application requiring a more optimized version of A/B testing. Adoption in robotics has been more limited, e.g., to selecting trajectories for object rearrangement planning \cite{Koval2015}, kicking strategies in robotic soccer \cite{Mendoza2016}, and, perhaps most closely related, selecting among deformable object models for acquisition tasks \cite{McConachie2017}. Unlike previous work, we argue that it is untenable to construct deformable object models for every food item, as conventional grocery stores typically stock in excess of 40,000 products \cite{Malito2017}. Instead, we take a model-free approach that operates directly on the image context space.

No-regret algorithms for solving bandit problems include UCB \cite{Auer2002} and EXP3 \cite{Auer2002a} for stochastic and adversarial reward distributions, respectively. They were also extended to the bandits-with-expert-advice setting (a generalization of the contextual bandit problem for small policy classes) with EXP4 \cite{Auer2002a}. Baseline methods for the contextual bandit problem include epoch-greedy \cite{Langford2008} and greedy \cite{Bastani2017}, both of which are simple to implement and perform well in practice, although they do not achieve optimal regret guarantees. More recent advances include LinUCB \cite{Li2010}, RegCB \cite{Foster2018} and Online Cover \cite{Agarwal2014}, a computationally efficient approximation to an algorithm that achieves optimal regret. For a recent and thorough overview, we refer the interested reader to \cite{Lattimore2019,Bietti2018}.

\subsection{Insights from Previous Work}

As noted previously, even when we control for covariate shift in a laboratory setting and switch to an unbiased success rate estimate (see \figref{fig:spanet}), SPANet is unable to generalize to some previously unseen food categories (specifically, kiwi and banana). We hypothesize that this lack of generalizability is due partly to the high diversity of actions for these food categories. For example, the most successful fork pitch for kiwi and banana was TA, which differs significantly from the successful actions for the rest of the food item data set. To determine whether collecting additional training data would solve this problem, we controlled for both the number of previously seen food classes and the total number of previously seen training examples. The results, shown in \figref{fig:data_amount}, do not noticeably improve out-of-class performance. An online learning approach lets  training continue indefinitely, bringing out-of-class food items into the effective training set. It also amortizes the potentially time-consuming process of data collection (SPANet's data set, for example, required approximately 81 hours of supervision) over the useful life of the system.

%%%%%%%%%%%%%%%%%%%%%%%%%%%%%%%%%%%%%%%%%%%%%%%%%%%%%%%%%%%%%%%
%%%%%%%%%%%%%%%%%

\section{ONLINE LEARNING WITH CONTEXTUAL BANDITS}
\subsection{Formulation}

A general contextual bandit algorithm consists of two parts: (1) an\textit{ exploration strategy} determines which action to take at each time step given the context and some policy, and (2) a \textit{learner} incorporates the bandit feedback received each time step into the policy. Algorithm \ref{alg:general} presents this structure as it applies in the environment with SPANet features.

At each round $t=1, \dots, T$, the interaction protocol consists of
\begin{enumerate}
	\item{\em Context observation.} The user selects a food item to acquire (in this work, we use RetinaNet \cite{2017Lin} to detect objects). We observe an RGBD image containing the single food item. We pass this through SPANet (Section \ref{sec:prev}) and use the penultimate layer as the context features $x_t \in \mathbb{R}^{d}$. The RGBD image is also used to localize the object for execution of the action.
	\item{\em Action selection.} The algorithm selects one manipulation strategy $a_t \in \mathcal{A} = \{1, 2, \dotsc, K\}$. In our initial implementation, $K=6$, with 3 pitch angles (VS, TV, TA) and 2 roll angles (parallel and perpendicular to the food), as shown in \figref{fig:overview}. The robot always skewers the center of the food item.
	\item{\em Partial loss observation.} The environment provides a binary loss $c_t(a_t, x_t) \in \{0, 1\}$, where $c_t = 0$ corresponds to the robot successfully acquiring the single desired food item.
\end{enumerate}
\figref{fig:system} presents a flow diagram of this protocol and its components.

The algorithm itself consists of a stochastic policy $\pi(x_t) = \mathbb{P}(a_t = a | x_t)$, and the goal is to minimize the cumulative regret of this policy. In other words, we wish to minimize $R_T$, which is the difference in performance between our policy $\pi$ and the best possible policy $\pi^*\in\Pi$ for the lifetime of our program $T$. With $c_t\in\mathcal{C}$, $x_t\in\mathcal{X}$, $a_t\in\mathcal{A}$ at time $t$, we have
\begin{align}
        R_T := &\sum_{t=1}^T c_t(\pi(\phi(x_t))) - \sum_{t=1}^T c_t(\pi^*(\phi(x_t))).
\end{align}

In cases where we compare algorithms with different sets $\Pi$, such as when tuning on dimension $d$ as a hyper-parameter, we instead try to minimize cumulative loss, the first term of $R_T$.

\begin{algorithm}[t]
\caption{General Contextual Bandit with SPANet Features}
\label{alg:general}
\textbf{Input:} Trained SPANet $\phi$, Environment $E$\\
\textbf{Initialize} Context $x \in \mathcal{X} \sim E$\\
\For{$t=1, \dotsc, T$}{
    Find features $\phi(x)$\\ 
    $p_t$ $\leftarrow$ {\tt explore}($\phi(x)$)\\
    Select action $a_t \sim p_t$\\
    Receive $c_t \sim E | a_t$\\
    {\tt learn}($\phi(x)$, $a_t$, $c_t$, $p_t$)\\
    \If{$c_t = 0$}{
        Re-sample context $x \sim E$
    }
}
\end{algorithm}
\subsection{Learning: Importance-Weighted Linear Regression}
\label{sec:iwlr}
The learning portion of a contextual bandit algorithm operates by first using past observations to estimate the cost of all actions for a given context. This reduces the problem to off-policy supervised learning. Since the contextual bandit literature tends to focus on exploration strategy, the sub-algorithm that performs the underlying full-feedback classification or regression is referred to as an {\em oracle}. All algorithms we define here use an importance-weighted linear regression oracle.

For our feature extractor, we use the activation of the penultimate layer in SPANet and fine tune the final layer in an online fashion. Thus, justified by the success of SPANet, we assume a linear map from the $\mathbb{R}^{d}$ features to the expected cost of each action: $\mathbb{E}[c_t|a_t,x_t] = \theta_{a_t}^T\phi(x_t)$. In this case, the regression oracle computes a weighted least-squares estimate
\begin{align}
    \widehat{\theta} := \sum_{t=0}^T\frac{1}{p_t(a_t)}\left(\theta_{a_t}^T\phi(x_t) - c_t(a_t)\right)^2.
\end{align}
Similarly to inverse propensity-scoring \cite{Agarwal2014}, the weight $p_t(a_t)^{-1}$ ensures that this returns an unbiased estimate of the underlying true weights $\theta^*$. An implementation of this oracle is shown in Algorithm \ref{alg:iwr}. The policy associated with a given weight estimate $\widehat{\theta}$ is the greedy policy: $\pi_{\theta}(x) = \arg\min_{a}\theta_{a}^T\phi(x)$.

\begin{algorithm}[t]
\caption{Importance-Weighted Regression Oracle}
\label{alg:iwr}
\SetKwFunction{FOracle}{learn}
\SetKwProg{Fn}{Function}{:}{}
\textbf{Input:} Regularization parameter $\lambda$, $d$ (features)\\
\textbf{Initialize} $\pi_0$: $\forall a \in \mathcal{A}$: $\mathbf{A}_a\leftarrow \lambda \mathbf{I}_{d\times d}; b_a\leftarrow 0$ \\
\SetKwProg{Pn}{Function}{:}{\KwRet}
\Pn{\FOracle{$\pi$, $\phi(x)$, $a_t$, $c_t$, $p_t(a_t)$}}{
    $(\mathbf{A}, b) \leftarrow \pi$\\
	$\mathbf{A}_{a_t} \leftarrow \mathbf{A}_{a_t} + \frac{1}{p_t}\phi\phi^T$ \\
	$b_{a_t} \leftarrow b_{a_t} + \frac{c_t}{p_t}\phi$ \\
	$\widehat{\theta}_{a_t} \leftarrow \mathbf{A}_{a_t}^{-1}b_{a_t}$ \\
	$\pi' \leftarrow (\widehat{\theta}, \mathbf{A}, b)$
}
\end{algorithm}
\begin{algorithm}[t]
\caption{$\epsilon$-greedy}
\label{alg:greedy}
\SetKwFunction{FExplore}{explore}
\SetKwProg{Fn}{Function}{:}{}
\textbf{Input:} Exploration parameter $\epsilon\in[0,1)$\\
\SetKwProg{Pn}{Function}{:}{\KwRet}
\Pn{\FExplore{$\phi(x)$}}{
	$p_t(a) \leftarrow \frac{\epsilon}{K} + (1-\epsilon)\1\{\pi_t(\phi(x))\}$
}
\end{algorithm}

\label{sec:conban}
\subsection{Exploration Strategy: $\epsilon$-greedy}

One of the simplest approaches to exploration is the $\epsilon$-greedy algorithm, shown in Algorithm \ref{alg:greedy}. This algorithm opts for the optimal action based on previous observations with probability $(1-\epsilon)$ and explores all actions uniformly with probability $\epsilon$. We consider both purely greedy ($\epsilon=0$) and exploratory ($\epsilon > 0$) variants.

With arbitrary contexts, the $\epsilon$-greedy algorithm (with optimized $\epsilon$) has a cumulative regret bound $R_T \sim O(T^{2/3})$, though it can perform well empirically \cite{Bietti2018}. Repeated contexts on failure also enables a better regret bound since taking multiple actions can provide effectively better-than-bandit feedback for a given context.

\subsection{Exploration Strategy: LinUCB}
\label{sec:linucb}
The other algorithm we use is Linear Upper Confidence bound (LinUCB), presented in Algorithm \ref{alg:single}. We justify the use of LinUCB \cite{Abbasi2011} due to the linear form of the ultimate SPANet layer (as justified in Section \ref{sec:iwlr}). Unlike $\epsilon$-greedy, the regret bound for LinUCB holds even if an adversary were choosing the worst-case contexts to show. Therefore, LinUCB can in theory be robust against covariate shift, allowing it to potentially be very competitive in this setting.

At each time step, we choose the action that maximizes the reward UCB (or, equivalently, loss LCB). This implicitly encourages exploration. In a choice between two actions with similar expected costs, the algorithm opts for the one with higher variance. With arbitrary contexts, LinUCB has a cumulative regret bound $R_T \sim O(T^{1/2})$, an improvement over $\epsilon$-greedy in the worst case. Like $\epsilon$-greedy, seeing repeated contexts on failure may improve this bound.

\begin{algorithm}[t]
\caption{LinUCB}
\label{alg:single}
\SetKwFunction{FExplore}{explore}
\SetKwProg{Fn}{Function}{:}{}
\textbf{Input:} Width parameter $\alpha$\\
\SetKwProg{Pn}{Function}{:}{\KwRet}
\Pn{\FExplore{$\phi(x)$}}{
    \For{$a\in\mathcal{A}$}{
	    $ucb_a \leftarrow \theta_a^T \phi(x)+ \alpha \sqrt{\phi(x)^T \mathbf{A}_a^{-1} \phi(x)} $\ \cite{Chu2011}\\
	}
}
\end{algorithm}

%%%%%%%%%%%%%%%%%%%%%%%%%%%%%%%%%%%%%%%%%%%%%%%%%%%%%%%%%%%%%%%%%%%%%%%%%%%%%%%%
\section{EXPERIMENTS}
\subsection{Tuning in Simulation}
\begin{figure*}[t!]
    \centering
    \includegraphics[width=0.32\linewidth]{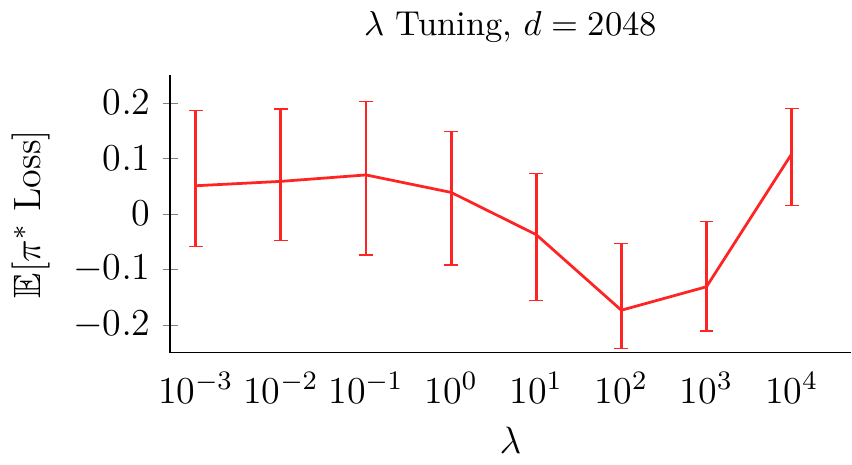}
    \hfill
    \includegraphics[width=0.32\linewidth]{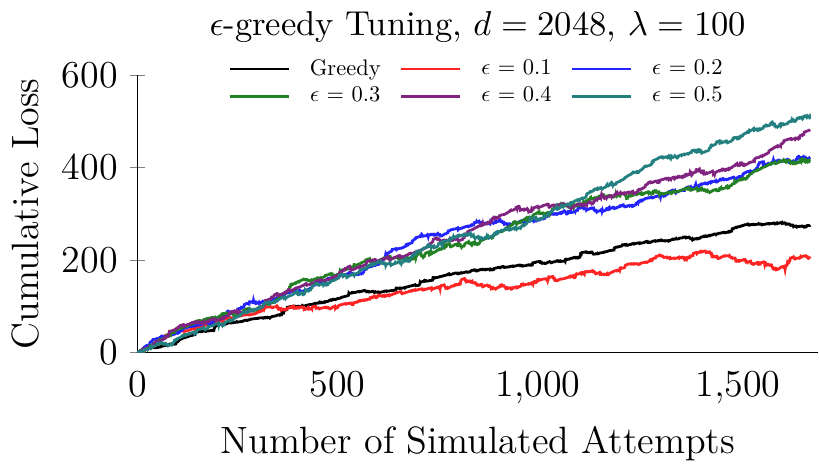}
    \includegraphics[width=0.32\linewidth]{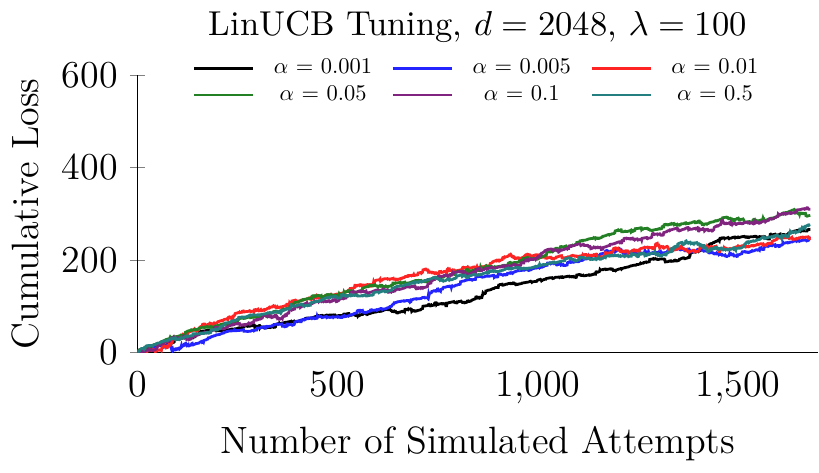}
    \caption{Hyper-parameter tuning in simulation, with banana, apple, and grapes excluded from SPANet. (\emph{Left}) $\pi^*$ performance as a function of $\lambda$ with $95\%$ confidence intervals. (\emph{Center/Right}) cumulative loss of the contextual bandit algorithms on the excluded food items as a function of their exploration hyper-parameters. On this data set, LinUCB exhibits more stable competitive performance than $\epsilon$-greedy.}
    \label{fig:simu}
    \vspace{-0.3cm}
\end{figure*}

\begin{figure*}[t!]
    \centering
    \includegraphics[width=0.25\linewidth]{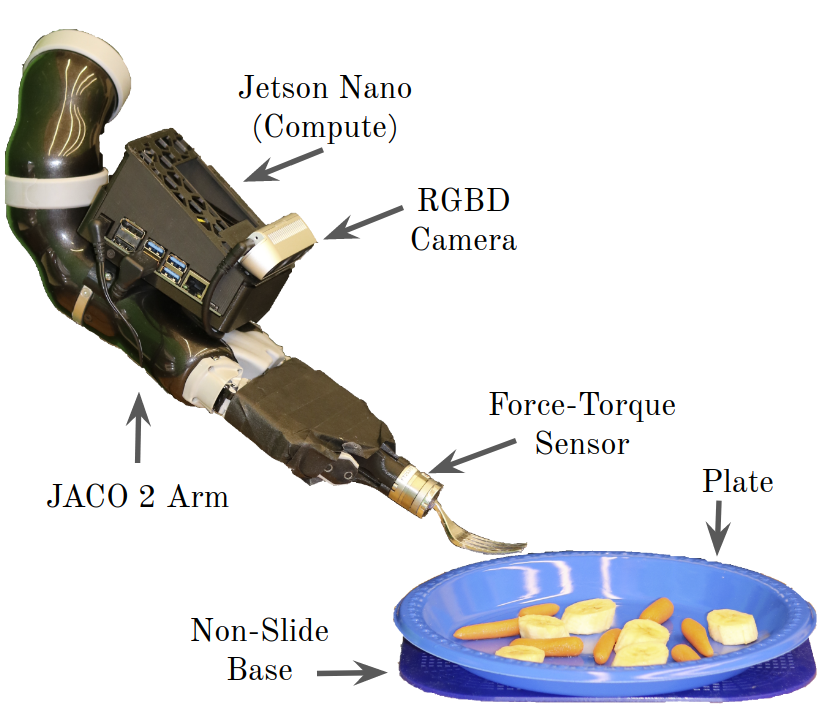}
    \hfill
    \includegraphics[width=0.73\linewidth]{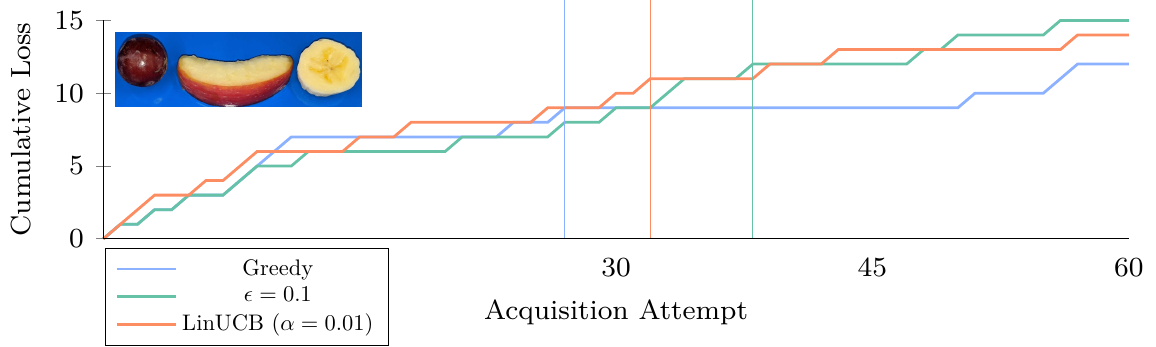}
    \caption{Results of Experiment 1 \emph{(Left)} using the Autonomous Dexterous Arm (ADA) \emph{(Right)}. SPANet was trained on 12 food types -- excluding apples, bananas, and grapes -- 3 types of food with significantly different success rate distributions over strategy. Initialized to $\theta_0=\vec{0}$, the robot cycled 20 times through all 3 food types. For each algorithm, the vertical line represents the point after which $100\%$ of the strategies selected were among the best strategies for the food item observed. Greedy is competitive since no pre-training weighs it down (as in Experiment 2), and $\epsilon$-greedy can still choose a poor strategy after convergence. Regardless, all algorithms converge within $\sim 10$ failures per food item.}
    \label{fig:exp1}
    \vspace{-0.5cm}
\end{figure*}

\label{sec:dr}

In addition to the normal hyper-parameters associated with linear regression (dimension $d$ and L2 regularization parameter $\lambda$), each algorithm has its own exploration hyper-parameter. We tune these by constructing a simulated training environment using the data from \cite{2019Feng}. Specifically, we exclude from SPANet three food items with very different success rate distributions over strategies. Banana slices are very sensitive to fork pitch, with TA performing the best by a wide margin because it prevents the slice from slipping off the fork. Grapes are, in general, very difficult to pick up, with the best strategy still dependent on biases in perception and planning. Apple slices are, in general, very easy to acquire, with some sensitivity to fork roll angle due to their length. Based on \cite{2019Feng}, VS, perpendicular roll angle, is likely the best strategy by a slight margin.

Since this data, by necessity, was collected with bandit feedback, the original work imputed the full loss vector of each context by averaging the success rate of a given action across all food items of the same type. While simple, this can introduce a herding bias into the simulation relative to the real world. We eliminate bias in our data set using a doubly robust \cite{Dudik2011} estimator
\begin{align}
\hat{l}_{DR}(x_i, a) = \hat{l}_a + (l_i - \hat{l}_a) \frac{\1(a_i = a)}{p(a_i|x_i)},
\end{align}
where $\hat{l_a}$ is the imputed value from herding, $p(a_i|x_i)$ is the probability that we took action $a_i$ during data collection ($\frac{1}{6}$ in our case since data was collected uniformly across all actions), and $l_i$ is the actual binary loss associated with that sample (only available for $a_i$). This estimator eliminates bias (i.e., $\E[\hat{l}_{DR}] = l$) from our imputed values at the cost of added variance. For each set of hyper-parameters, $\pi^*$ is determined by performing full-feedback least-squares linear regression on all previously unseen food items to estimate $\theta^*$.

First, we tuned the linear regression parameters $d$ and $\lambda$. Using the original SPANet feature space of $\mathbb{R}^{2048}$, we found that we needed significant regularization (large $\lambda$) to see any results on our limited data set. However, while reducing our feature-space dimension $d$ could in theory improve our regret bounds (e.g., LinUCB's $R_T \sim O(d)$), it empirically reduced our best possible ($\pi^*$) performance. This exposed us to a two-dimensional trade-off of bias vs. variance and performance vs. data efficiency. For $d$, the hit to $\pi^*$ outweighs any improvements in regret. For $\lambda$, while $100$ and $1000$ produced similar $\pi^*$ performance (as shown in \figref{fig:simu}a), $\lambda=100$ performed better on greedy cumulative loss.

Figure \ref{fig:simu}b,c show the results of tuning the exploration parameters $\epsilon$ and $\alpha$. Note that a greater loss is expected since the doubly robust losses have a higher variance than reality. Stochastic $\epsilon$-greedy showed a clear local minimum at $\epsilon=0.1$. Meanwhile, LinUCB demonstrated more consistent competitive performance across multiple orders of magnitude for $\alpha$. We selected $\alpha=0.01$, which reached a slight minimum loss, for the real robot experiments.
\begin{figure*}[t!]
    \centering
    \includegraphics[width=0.32\linewidth]{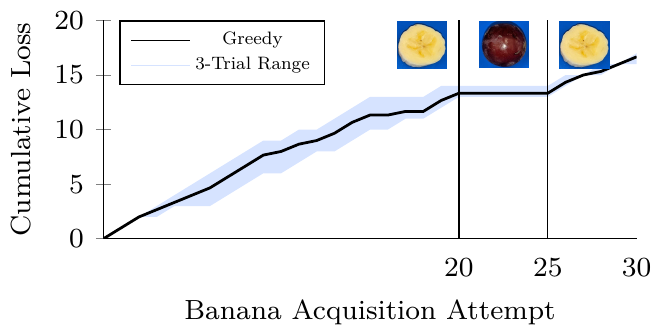}
    \includegraphics[width=0.32\linewidth]{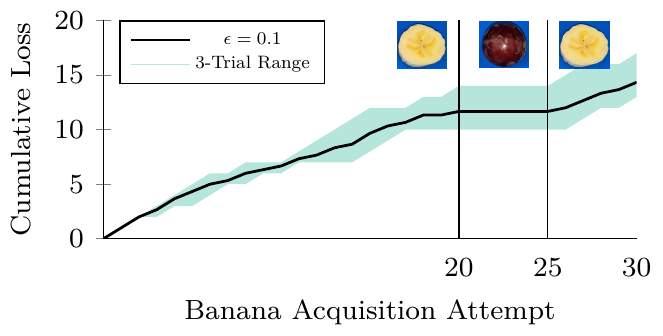}
    \includegraphics[width=0.32\linewidth]{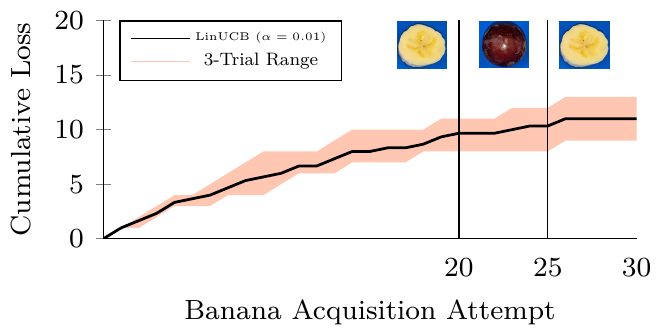}
    \includegraphics[width=0.32\linewidth]{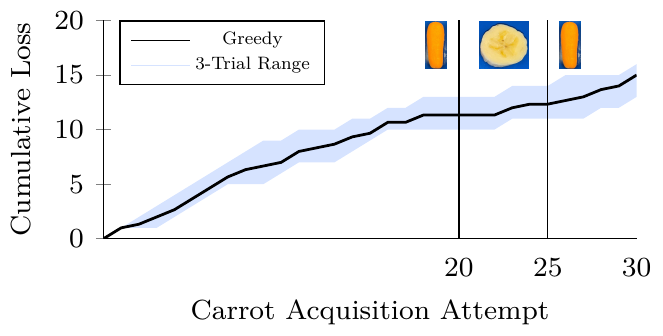}
    \includegraphics[width=0.32\linewidth]{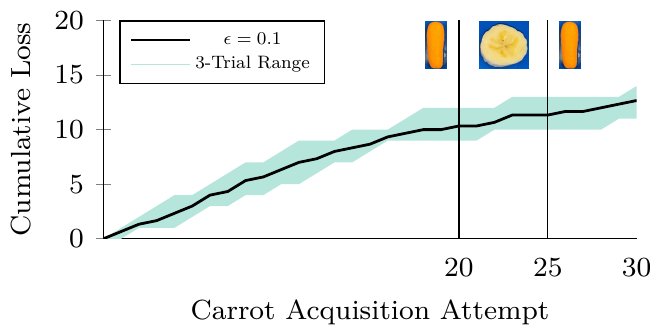}
    \includegraphics[width=0.32\linewidth]{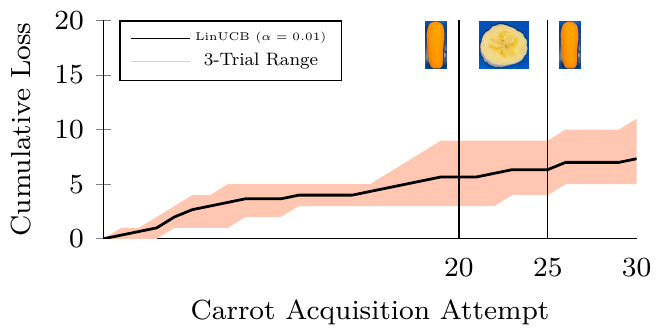}
    \vspace{-0.3cm}
    \caption{Empirical cumulative loss for each contextual bandit algorithm on an unseen food item when $\theta_0$ was trained on all previously seen food items. For each trial, attempts 20-25 were performed on a different, previously seen food item to demonstrate that the policy did not forget its best strategy during the unseen food attempts. LinUCB demonstrates the most stable performance, especially when there are multiple good strategies (as for carrots).}
    \label{fig:exp2}
    \vspace{-0.3cm}
\end{figure*}
\begin{figure*}[t!]
    \centering
    \includegraphics[width=0.26\linewidth]{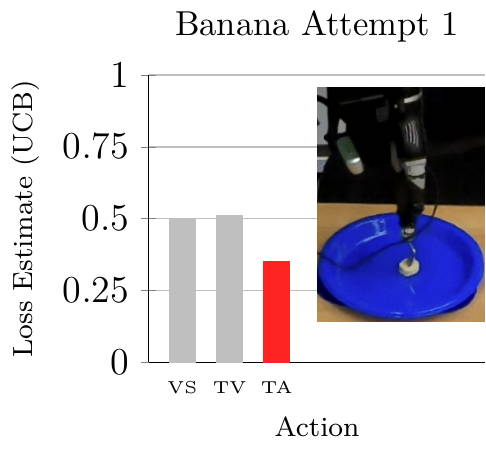}
    \includegraphics[width=0.23\linewidth]{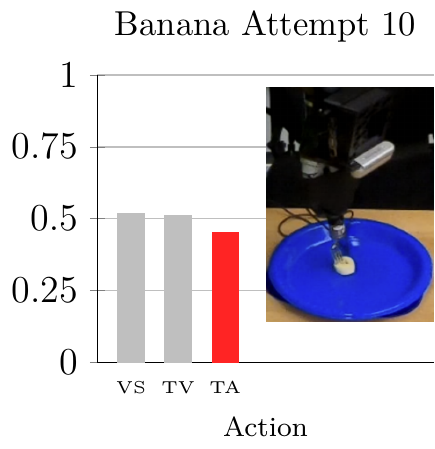}
    \hfill
    \includegraphics[width=0.23\linewidth]{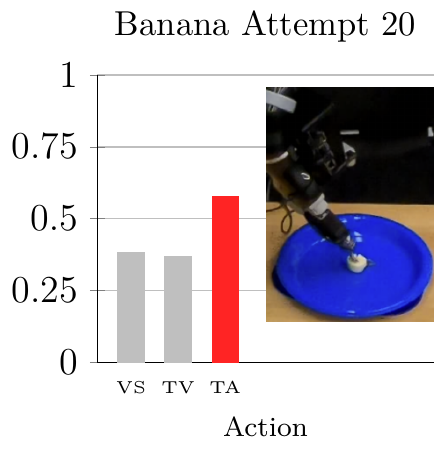}
    \includegraphics[width=0.23\linewidth]{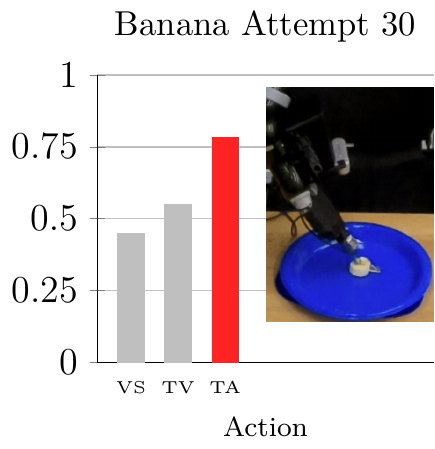}
    \vspace{-0.3cm}
    
    \caption{Evolution of the internal upper confidence bound (UCB) estimate of the success rate of each action over time for one of the banana experiments. Since banana slices have rotational symmetry, for each fork pitch we present only the maximum UCB of the two roll angles. The optimal fork pitch for acquiring banana, tilted-angled (TA), is highlighted in red and increases over the course of the experiment.}
    \label{fig:internal}
    \vspace{-0.6cm}
\end{figure*}

\subsection{Real Robot Experiments}

\paragraph{System description.}
Our setup, the Autonomous Dexterous Arm (ADA) (\figref{fig:exp1}, left), consists of a 6 DoF JACO robotic arm \cite{Jaco2018}. The arm has 2 fingers that grab an instrumented fork (forque) using a custom-built, 3D-printed fork holder. The system uses visual and haptic modalities to perform the feeding task. For haptic input, we instrumented the forque with a 6-axis ATI Nano25 Force-Torque sensor \cite{Forque2018}. We use haptic sensing to control the end effector forces during skewering. Specifically, force thresholds are used as hard-coded transition cues between motion primitives. For visual input, we mounted a custom built wireless perception unit on the robot’s wrist; the unit includes the Intel RealSense D415 RGBD camera and the NVidia Jetson Nano for wireless transmission. Food is placed on a plate mounted on an anti-slip mat commonly found in assisted living facilities.

\paragraph{General procedure.}

For each attempt, we place a single food item in the center of the plate. ADA positions itself vertically above the plate and performs object detection and featurization using a checkpoint of SPANet that was trained with some food items excluded. Importantly, the identity of the food items, while used for object detection, was {\em never} made available to the contextual bandit algorithm. After performing the requested action, the binary loss is recorded manually, and the learning algorithm is updated. To mimic a realistic feeding setting, we removed and replaced the food item only after a successful acquisition.

We define a bite acquisition attempt as a success $(c_t=0)$ if the target food item, either the whole piece or a cut portion, remains on the fork for $5$ seconds after removal from the plate. If the target food item is skewered with at least 2 out of 4 tines but the fork fails to pick it up or the food falls off soon after lift-off, the attempt is deemed a failure $(c_t=1)$. If less than 2 out of 4 tines touch a food item due to system-level errors (e.g., perception or planning), we discard the attempt completely.

\paragraph{Experiment 1.}

This experiment tests whether the features generated by SPANet trained without previously unseen food items are rich enough for the contextual bandit algorithm to find the best strategy for multiple food items. We cycle through 3 food items (apple, banana, then grape) 20 times, leading to 60 total attempts. We choose these items for the same reason as the simulation: they are representative of the majority of our food data set.

\paragraph{Experiment 2.}

This experiment tests whether the contextual bandit algorithms can adapt to new food items when given a $\theta$ that has already been trained on many previously seen dissimilar food items from the doubly robust simulated environment. Unlike Experiment 1, we test on only one food item at a time, so the set of dissimilar food items is of a non-negligible size. For banana slices, $\theta$ was trained on all $\sim 8000$ attempts on all 15 non-banana food items because it is the only food item sensitive to fork pitch, where TA is the best strategy. For carrots, which are very sensitive to fork roll (i.e., VS and TV, perpendicular roll angle, are the likely best strategies by a wide margin), $\theta$ was trained on $\sim 3000$ attempts, which excluded other food items sensitive to fork roll, such as apples, bell peppers, and celery. For each food item, we conducted 20 attempts, followed by 5 attempts with a previously seen food item (grape and banana, respectively), followed by another 5 attempts of the test food item, to ensure that $\pi$ did not forget previously seen food items after adapting to a new one.

%%%%%%%%%%%%%%%%%%%%%%%%%%%%%%%%%%%%%%%%%%%%%%%%%%%%%%%%%%%%%%%%%%%%%%%%%%%%%%%%

\section{RESULTS}
\figref{fig:exp1} (right) summarizes the results of Experiment 1. All algorithms suffered a cumulative loss between 10 and 15. The key takeaway is that all algorithms converged to the best strategy set within $\sim 10$ failures per new food item, after which the best strategy (or a strategy within the best set of strategies) was chosen $100\%$ of the time for each food item. The only subsequent errors were due to uncertainties in perception and planning. Interestingly, greedy had the highest performance using this metric, though, unlike Experiment 2, it was not weighed down by pretraining in $\theta$, and greedy is often empirically competitive in contextual bandit settings \cite{Bietti2018}. These results suggest that the SPANet features are indeed rich enough for contextual bandit algorithms to learn the best strategy for multiple representative food items simultaneously.

\figref{fig:exp2} summarizes the results of Experiment 2. LinUCB exhibited superior cumulative loss performance for both food items, and greedy exhibited particularly poor performance. $\epsilon$-greedy produced higher-variance results, spanning from the best performance of greedy to the worst performance of LinUCB. The inverse of Experiment 1, this is probably due to the weight of the pretrained $\theta$ forcing greedy to try previously good strategies before exploring new ones. LinUCB could capitalize on the uncertainty introduced by seeing a significantly different context. \figref{fig:internal} shows how LinUCB's upper confidence bound estimates changed over time as it adapted to bananas. Regardless, its consistent performance on the previously seen food item did demonstrate that the contextual bandit algorithm could adapt to new information without forgetting the best strategies for previously seen food items.

In general, it is difficult to map an online learning metric like regret to a static metric like acquisition success rate. That said, as regret approaches 0, we expect that our framework will approach the success rate of the fully trained SPANet, cited in \protect\cite{2019Feng} as approximately 75\%. Both experiments suggest that this convergence could happen within 10 attempts, even if the previously unseen food requires a completely new acquisition strategy.

%%%%%%%%%%%%%%%%%%%%%%%%%%%%%%%%%%%%%%%%%%%%%%%%%%%%%%%%%%%%%%%%%%%%%%%%%%%%%%%%
\section{DISCUSSION}
One key takeaway from these results is that LinUCB is empirically robust across a range of hyper-parameters and initial conditions. A fluke early failure will not sink a high-expectation action since the increasing variance dampens the decreasing expectation. Robustness is vital for a robotic feeding system: users, especially those with some mobility, may not tolerate too many errors in an autonomous system they use daily \cite{bhattacharjee2020userpref}. While the number of failures seen here may not be acceptable for a single meal, both experiments suggest that this is a 1-time cost that can be amortized over the life of the feeding system.

In future work, we intend to broaden our scope to multiple food items by considering the entire plate of food items as a single compound state, or just switching food items if the expected success rate of all actions falls below some threshold.

Beyond using RGBD context features, our robot has access to other modalities, including haptic feedback. Non-destructive probing can provide us a richer context, especially if we need to differentiate between similar-looking food items with different material properties (say, because one is cooked or ripe). Other groups have found success using a vibration-detecting audio modality \cite{Clarke2018} as well.

Finally, we investigated only discrete, solid food items. To generalize to a realistic average plate with continuous and mixed foods, we will need to expand to a richer action space. Since adding more action parameters (e.g.\ yaw, where on the food item to skewer, skewering force) will increase the size of the action space at a combinatorial rate, we could leverage similarities between actions by modeling each one as a coupled slate of actions \cite{Dimakopoulou2019}.

Overall, these results suggest that a contextual bandit approach with discrete, dissimilar actions offers a promising route to data-efficient adaptive bite acquisition.

\section*{Acknowledgments}
This work was partially funded by the National Institute of Health R01 (\#R01EB019335), National Science Foundation CPS (\#1544797), National Science Foundation NRI (\#1637748), the Office of Naval Research, the RCTA, Amazon, and Honda Research Institute USA. We would also like to thank Jaclyn Brockschmidt for her help with the robot experiments.

\addtolength{\textheight}{-6cm}
\begin{spacing}{0.89}
\bibliographystyle{IEEEtran}
\bibliography{IEEEabrv,main}

\begin{thebibliography}{10}
\providecommand{\url}[1]{#1}
\csname url@rmstyle\endcsname
\providecommand{\newblock}{\relax}
\providecommand{\bibinfo}[2]{#2}
\providecommand\BIBentrySTDinterwordspacing{\spaceskip=0pt\relax}
\providecommand\BIBentryALTinterwordstretchfactor{4}
\providecommand\BIBentryALTinterwordspacing{\spaceskip=\fontdimen2\font plus
\BIBentryALTinterwordstretchfactor\fontdimen3\font minus
  \fontdimen4\font\relax}
\providecommand\BIBforeignlanguage[2]{{%
\expandafter\ifx\csname l@#1\endcsname\relax
\typeout{** WARNING: IEEEtran.bst: No hyphenation pattern has been}%
\typeout{** loaded for the language `#1'. Using the pattern for}%
\typeout{** the default language instead.}%
\else
\language=\csname l@#1\endcsname
\fi
#2}}

\bibitem{2012Brault}
M.~W. Brault, ``Americans with disabilities: 2010,'' Current population
  reports, vol. 7, pp. 70–131, 2012.

\bibitem{1990Prior}
S.~D. Prior, ``An electric wheelchair mounted robotic arm-a survey of potential
  users,'' Journal of medical engineering \& technology, vol. 14, no. 4, pp.
  143–154, 1990.

\bibitem{1994Stanger}
C.~A. Stanger, C.~Anglin, W.~S. Harwin, and D.~P. Romilly, ``Devices for
  assisting manipulation: a summary of user task priorities,'' IEEE
  Transactions on rehabilitation Engineering, vol. 2, no. 4, pp. 256–265,
  1994.

\bibitem{myspoon}
MySpoon, 2018, \url{https://www.secom.co.jp/english/myspoon/food.html}.

\bibitem{obi}
Obi, 2018, \url{https://meetobi.com/}.

\bibitem{bhattacharjee2018food}
T.~Bhattacharjee, G.~Lee, H.~Song, and S.~S. Srinivasa, ``Towards robotic
  feeding: Role of haptics in fork-based food manipulation,'' \emph{IEEE
  Robotics and Automation Letters}, 2019.

\bibitem{2019Feng}
R.~Feng, Y.~Kim, G.~Lee, E.~K. Gordon, M.~Schmittle, S.~Kumar,
  T.~Bhattacharjee, and S.~S. Srinivasa, ``Robot-assisted feeding: Generalizing
  skewering strategies across food items on a realistic plate,'' in
  \emph{International Symposium on Robotics Research}, 2019.

\bibitem{2019Gallenberger}
D.~Gallenberger, T.~Bhattacharjee, Y.~Kim, and S.~Srinivasa, ``Transfer depends
  on acquisition: Analyzing manipulation strategies for robotic feeding,''
  ACM/IEEE International Conference on Human-Robot Interaction, 2019.

\bibitem{Bietti2018}
A.~Bietti, A.~Agarwal, and J.~Langford, ``A contextual bandit bake-off,''
  \emph{arXiv preprint 1802.04064}, Feb. 2018.

\bibitem{Chu2011}
C.~Wei, L.~Li, L.~Reyzin, and R.~E. Schapire, ``Contextual bandits with linear
  payoff functions,'' in \emph{Proceedings of the Fourteenth International
  Conference on Artificial Intelligence and Statistics}, 2011, pp. 208--214.

\bibitem{chua2003robotic}
P.~Chua, T.~Ilschner, and D.~Caldwell, ``Robotic manipulation of food
  products--a review,'' \emph{Industrial Robot: An International Journal},
  vol.~30, no.~4, pp. 345--354, 2003.

\bibitem{erzincanli1997meeting}
F.~Erzincanli and J.~Sharp, ``Meeting the need for robotic handling of food
  products,'' \emph{Food Control}, vol.~8, no.~4, pp. 185--190, 1997.

\bibitem{morales2014soft}
R.~Morales, F.~Badesa, N.~Garcia-Aracil, J.~Sabater, and L.~Zollo, ``Soft
  robotic manipulation of onions and artichokes in the food industry,''
  \emph{Advances in Mechanical Engineering}, vol.~6, p. 345291, 2014.

\bibitem{brett1991research}
P.~Brett, A.~Shacklock, and K.~Khodabendehloo, ``Research towards generalised
  robotic systems for handling non-rigid products,'' in \emph{International
  Conference on Advanced Robotics}.\hskip 1em plus 0.5em minus 0.4em\relax
  IEEE, 1991, pp. 1530--1533.

\bibitem{williams2001teaching}
T.~G. Williams, J.~J. Rowland, and M.~H. Lee, ``Teaching from examples in
  assembly and manipulation of snack food ingredients by robot,'' in
  \emph{{IEEE/RSJ} International Conference on Intelligent Robots and Systems},
  vol.~4.\hskip 1em plus 0.5em minus 0.4em\relax IEEE, 2001, pp. 2300--2305.

\bibitem{blanes2011technologies}
C.~Blanes, M.~Mellado, C.~Ortiz, and A.~Valera, ``Technologies for robot
  grippers in pick and place operations for fresh fruits and vegetables,''
  \emph{Spanish Journal of Agricultural Research}, vol.~9, no.~4, pp.
  1130--1141, 2011.

\bibitem{brosnan2002inspection}
T.~Brosnan and D.-W. Sun, ``Inspection and grading of agricultural and food
  products by computer vision systems--a review,'' \emph{Computers and
  Electronics in Agriculture}, vol.~36, no.~2, pp. 193--213, 2002.

\bibitem{du2006learning}
C.-J. Du and D.-W. Sun, ``Learning techniques used in computer vision for food
  quality evaluation: a review,'' \emph{Journal of food engineering}, vol.~72,
  no.~1, pp. 39--55, 2006.

\bibitem{ding1994shape}
K.~Ding and S.~Gunasekaran, ``Shape feature extraction and classification of
  food material using computer vision,'' \emph{Transactions of the ASAE},
  vol.~37, no.~5, pp. 1537--1545, 1994.

\bibitem{ma2011chinese}
W.-T. Ma, W.-X. Yan, Z.~Fu, and Y.-Z. Zhao, ``A chinese cooking robot for
  elderly and disabled people,'' \emph{Robotica}, vol.~29, no.~6, pp. 843--852,
  2011.

\bibitem{sugiura2010cooking}
Y.~Sugiura, D.~Sakamoto, A.~Withana, M.~Inami, and T.~Igarashi, ``Cooking with
  robots: designing a household system working in open environments,'' in
  \emph{Proceedings of the SIGCHI Conference on Human Factors in Computing
  Systems}.\hskip 1em plus 0.5em minus 0.4em\relax ACM, 2010, pp. 2427--2430.

\bibitem{bollini2011bakebot}
M.~Bollini, J.~Barry, and D.~Rus, ``Bakebot: Baking cookies with the pr2,'' in
  \emph{The PR2 workshop: results, challenges and lessons learned in advancing
  robots with a common platform, IROS}, 2011.

\bibitem{beetz2011robotic}
M.~Beetz, U.~Klank, I.~Kresse, A.~Maldonado, L.~M{\"o}senlechner, D.~Pangercic,
  T.~R{\"u}hr, and M.~Tenorth, ``Robotic roommates making pancakes,'' in
  \emph{{IEEE-RAS} International Conference on Humanoid Robots}.\hskip 1em plus
  0.5em minus 0.4em\relax IEEE, 2011, pp. 529--536.

\bibitem{oreovideo}
``Oreo separator machines,'' \url{https://vimeo.com/63347829},[Online;
  Retrieved on 1st February, 2018].

\bibitem{gemici2014learning}
M.~C. Gemici and A.~Saxena, ``Learning haptic representation for manipulating
  deformable food objects,'' in \emph{{IEEE/RSJ} International Conference on
  Intelligent Robots and Systems}.\hskip 1em plus 0.5em minus 0.4em\relax IEEE,
  2014, pp. 638--645.

\bibitem{2016Park}
D.~Park, Y.~K. Kim, Z.~M. Erickson, and C.~C. Kemp, ``Towards assistive feeding
  with a {General-Purpose} mobile manipulator,'' \emph{arXiv preprint
  arXiv:1605.07996}, May 2016.

\bibitem{herlant_thesis}
L.~V. Herlant, ``{Algorithms, Implementation, and Studies on Eating with a
  Shared Control Robot Arm},'' Ph.D. dissertation, The Robotics Institute
  Carnegie Mellon University, 2016.

\bibitem{Tang2013}
L.~Tang, R.~Rosales, A.~Singh, and D.~Agarwal, ``{Automatic ad format selection
  via contextual bandits},'' in \emph{Proceedings of the 22nd ACM international
  conference on Information {\&} Knowledge Management}.\hskip 1em plus 0.5em
  minus 0.4em\relax ACM, 2013, pp. 1587--1594.

\bibitem{Bottou2013}
L.~Bottou, J.~Peters, J.~Qui{\~{n}}onero-Candela, D.~X. Charles, D.~M.
  Chickering, E.~Portugaly, D.~Ray, P.~Simard, and E.~Snelson,
  ``{Counterfactual reasoning and learning systems: The example of
  computational advertising},'' \emph{The Journal of Machine Learning
  Research}, vol.~14, no.~1, pp. 3207--3260, 2013.

\bibitem{Klasnja2015}
P.~Klasnja, E.~B. Hekler, S.~Shiffman, A.~Boruvka, D.~Almirall, A.~Tewari, and
  S.~A. Murphy, ``{Microrandomized trials: An experimental design for
  developing just-in-time adaptive interventions.}'' \emph{Health Psychology},
  vol.~34, no.~S, p. 1220, 2015.

\bibitem{Hochberg2016}
I.~Hochberg, G.~Feraru, M.~Kozdoba, S.~Mannor, M.~Tennenholtz, and E.~Yom-Tov,
  ``{Encouraging physical activity in patients with diabetes through automatic
  personalized feedback via reinforcement learning improves glycemic
  control},'' \emph{Diabetes care}, vol.~39, no.~4, pp. e59--e60, 2016.

\bibitem{Shortreed2011}
S.~M. Shortreed, E.~Laber, D.~J. Lizotte, T.~S. Stroup, J.~Pineau, and S.~A.
  Murphy, ``{Informing sequential clinical decision-making through
  reinforcement learning: an empirical study},'' \emph{Machine learning},
  vol.~84, no. 1-2, pp. 109--136, 2011.

\bibitem{Awerbuch2004}
B.~Awerbuch and R.~D. Kleinberg, ``{Adaptive routing with end-to-end feedback:
  Distributed learning and geometric approaches},'' in \emph{Proceedings of the
  thirty-sixth annual ACM symposium on Theory of computing}.\hskip 1em plus
  0.5em minus 0.4em\relax ACM, 2004, pp. 45--53.

\bibitem{Mandel2014}
T.~Mandel, Y.-E. Liu, S.~Levine, E.~Brunskill, and Z.~Popovic, ``{Offline
  policy evaluation across representations with applications to educational
  games},'' in \emph{Proceedings of the 2014 international conference on
  Autonomous agents and multi-agent systems}.\hskip 1em plus 0.5em minus
  0.4em\relax International Foundation for Autonomous Agents and Multiagent
  Systems, 2014, pp. 1077--1084.

\bibitem{Wang2014}
X.~Wang, Y.~Wang, D.~Hsu, and Y.~Wang, ``{Exploration in interactive
  personalized music recommendation: a reinforcement learning approach},''
  \emph{ACM Transactions on Multimedia Computing, Communications, and
  Applications (TOMM)}, vol.~11, no.~1, p.~7, 2014.

\bibitem{Shen2015}
W.~Shen, J.~Wang, Y.-G. Jiang, and H.~Zha, ``{Portfolio choices with orthogonal
  bandit learning},'' in \emph{Twenty-Fourth International Joint Conference on
  Artificial Intelligence}, 2015.

\bibitem{Koval2015}
M.~C. Koval, J.~E. King, N.~S. Pollard, and S.~S. Srinivasa, ``{Robust
  trajectory selection for rearrangement planning as a multi-armed bandit
  problem},'' in \emph{2015 IEEE/RSJ International Conference on Intelligent
  Robots and Systems (IROS)}.\hskip 1em plus 0.5em minus 0.4em\relax IEEE,
  2015, pp. 2678--2685.

\bibitem{Mendoza2016}
J.~P. Mendoza, R.~Simmons, and M.~Veloso, ``{Online learning of robot soccer
  free kick plans using a bandit approach},'' in \emph{Twenty-Sixth
  International Conference on Automated Planning and Scheduling}, 2016.

\bibitem{McConachie2017}
D.~McConachie and D.~Berenson, ``{Bandit-based model selection for deformable
  object manipulation},'' \emph{arXiv preprint arXiv:1703.10254}, 2017.

\bibitem{Malito2017}
A.~Malito, ``{Grocery stores carry 40,000 more items than they did in the
  1990s},'' 2017.

\bibitem{Auer2002}
P.~Auer, N.~Cesa-Bianchi, and P.~Fischer, ``{Finite-time analysis of the
  multiarmed bandit problem},'' \emph{Machine learning}, vol.~47, no. 2-3, pp.
  235--256, 2002.

\bibitem{Auer2002a}
P.~Auer, N.~Cesa-Bianchi, Y.~Freund, and R.~E. Schapire, ``{The nonstochastic
  multiarmed bandit problem},'' \emph{SIAM journal on computing}, vol.~32,
  no.~1, pp. 48--77, 2002.

\bibitem{Langford2008}
J.~Langford and T.~Zhang, ``{The epoch-greedy algorithm for multi-armed bandits
  with side information},'' in \emph{Advances in neural information processing
  systems}, 2008, pp. 817--824.

\bibitem{Bastani2017}
H.~Bastani, M.~Bayati, and K.~Khosravi, ``{Mostly exploration-free algorithms
  for contextual bandits},'' \emph{arXiv preprint arXiv:1704.09011}, 2017.

\bibitem{Li2010}
L.~Li, W.~Chu, J.~Langford, and R.~E. Schapire, ``A {Contextual-Bandit}
  approach to personalized news article recommendation,'' \emph{arXiv preprint
  arXiv:1003.0146}, Feb. 2010.

\bibitem{Foster2018}
D.~J. Foster, A.~Agarwal, M.~Dud{\'\i}k, H.~Luo, and R.~E. Schapire,
  ``Practical contextual bandits with regression oracles,'' \emph{arXiv
  preprint arXiv:1803.01088}, Mar. 2018.

\bibitem{Agarwal2014}
A.~Agarwal, D.~Hsu, S.~Kale, J.~Langford, L.~Li, and R.~E. Schapire, ``Taming
  the monster: A fast and simple algorithm for contextual bandits,''
  \emph{arXiv preprint arXiv:1402.0555}, Feb. 2014.

\bibitem{Lattimore2019}
T.~Lattimore and C.~Szepesvari, \emph{{Bandit Algorithms}}.\hskip 1em plus
  0.5em minus 0.4em\relax Cambridge University Press, 2019.

\bibitem{2017Lin}
T.~Lin, P.~Goyal, R.~Girshick, K.~He, and P.~Doll{\'a}r, ``Focal loss for dense
  object detection,'' in \emph{2017 {IEEE} International Conference on Computer
  Vision ({ICCV})}, Oct. 2017, pp. 2999--3007.

\bibitem{Abbasi2011}
Y.~Abbasi-yadkori, D.~P{\'a}l, and C.~Szepesv{\'a}ri, ``Improved algorithms for
  linear stochastic bandits,'' in \emph{Advances in Neural Information
  Processing Systems 24}, J.~Shawe-Taylor, R.~S. Zemel, P.~L. Bartlett,
  F.~Pereira, and K.~Q. Weinberger, Eds.\hskip 1em plus 0.5em minus 0.4em\relax
  Curran Associates, Inc., 2011, pp. 2312--2320.

\bibitem{Dudik2011}
M.~Dudik, J.~Langford, and L.~Li, ``Doubly robust policy evaluation and
  learning,'' \emph{arXiv preprint arXiv:1103.4601}, Mar. 2011.

\bibitem{Jaco2018}
K.~JACO, 2018,
  \url{https://www.kinovarobotics.com/en/products/robotic-armseries}.

\bibitem{Forque2018}
A.-I. F.-T. Sensor, 2018,
  \url{https://www.ati-ia.com/products/ft/ft\_models.aspx?id=Nano25}.

\bibitem{bhattacharjee2020userpref}
T.~Bhattacharjee, E.~Gordon, R.~Scalise, M.~Cabrera, A.~Caspi, M.~Cakmak, and
  S.~Srinivasa, ``Is more autonomy always better? exploring preferences of
  users with mobility impairments in robot-assisted feeding,'' in
  \emph{{ACM/IEEE} International Conference on Human-Robot Interaction}, 2020.

\bibitem{Clarke2018}
S.~Clarke, T.~Rhodes, C.~G. Atkeson, and O.~Kroemer, ``Learning audio feedback
  for estimating amount and flow of granular material,'' in \emph{Proceedings
  of The 2nd Conference on Robot Learning}, ser. Proceedings of Machine
  Learning Research, A.~Billard, A.~Dragan, J.~Peters, and J.~Morimoto, Eds.,
  vol.~87.\hskip 1em plus 0.5em minus 0.4em\relax PMLR, 2018, pp. 529--550.

\bibitem{Dimakopoulou2019}
M.~Dimakopoulou, N.~Vlassis, and T.~Jebara, ``Marginal posterior sampling for
  slate bandits,'' in \emph{Proceedings of the Twenty-Eighth International
  Joint Conference on Artificial Intelligence}, 2019, pp. 2223--2229.

\end{thebibliography}
\end{spacing}
\end{document}